\documentclass[sigconf]{acmart}
\AtBeginDocument{%
  }

\setcopyright{acmlicensed}
\copyrightyear{2018}
\acmYear{2018}
\acmDOI{XXXXXXX.XXXXXXX}
\acmConference[Conference acronym 'XX]{Make sure to enter the correct
  conference title from your rights confirmation email}{June 03--05,
  2018}{Woodstock, NY}
\acmISBN{978-1-4503-XXXX-X/2018/06}

\usepackage{color, colortbl}
\usepackage[utf8]{inputenc} 
\usepackage[T1]{fontenc}    
\usepackage{url}

\usepackage{hyperref}
\usepackage{wrapfig}
\usepackage{ulem}
\usepackage{pifont}
\usepackage{mathtools}
\usepackage{multirow}
\usepackage{makecell}
\usepackage{booktabs}
\usepackage{algorithm}
\usepackage{algpseudocode}
\usepackage{algorithmicx}
\usepackage[OT1]{fontenc}
\usepackage{enumitem}
\usepackage{cuted}
\usepackage{colortbl}

\definecolor{myorange}{HTML}{F37021}

\begin{document}

\title{\texttt{QuantMoE-Bench}: Examining Post-Training Quantization for Mixture-of-Experts}


\author{Pingzhi Li$^\dagger$}
\affiliation{%
  \institution{UNC-Chapel Hill}
  \city{Chapel Hill}
  \state{NC}
  \country{US}}
\email{pingzhi@cs.unc.edu}

\author{Xiaolong Jin$^\dagger$}
\affiliation{%
 \institution{Purdue University}
 \city{West Lafayette}
 \state{IN}
 \country{US}}
\email{jinxiaolong1129@gmail.com}

\author{Zhen Tan}
\affiliation{%
  \institution{Arizona State University}
  \city{Tempe}
  \state{AZ}
  \country{US}
}
\email{ztan36@asu.edu}

\author{Yu Cheng}
\affiliation{%
 \institution{The Chinese University of Hong Kong}
 \country{Hong Kong}}
\email{chengyu@cse.cuhk.edu.hk}

\author{Tianlong Chen}
\affiliation{%
  \institution{UNC-Chapel Hill}
  \city{Chapel Hill}
  \state{NC}
  \country{US}}
\email{tianlong@cs.unc.edu}

\renewcommand{\shortauthors}{Li et al.}

\begin{abstract}
Mixture-of-Experts (MoE) is a promising way to scale up the
learning capacity of large language models. It increases the number of parameters while keeping FLOPs nearly constant during inference through sparse activation.
Yet, it still suffers from significant memory overheads due to the vast parameter size, necessitating model compression techniques. 
Post-training quantization offers a powerful approach for model compression. Existing methods adopt a fixed quantization precision for the entire MoE model. This rigid setup can lead to suboptimal performance, without considering the inherent sparse structure.
For example, MoE's sparse routing mechanism leads to different activation patterns, where shared experts are accessed by all tokens while token-conditioned experts are selectively activated. This activation disparity suggests different quantization requirements, with consistently activated shared experts potentially needing higher precision to maintain model quality.
In this paper, we study a fine-grained precision setup for MoE quantization. We explore MoE structure-aware quantization heuristics, ranging from coarse (\textit{e.g.}, MoE layers) to fine granularity (\textit{e.g.}, linear layers). 
Our investigations reveal critical principles, where different MoE structures require \textit{varying numbers of bits} for effective quantization. Conclusions are supported by extensive benchmarking across two representative MoE models and six tasks including commonsense reasoning and natural language understanding. We further show that an MoE quantized in a fined-grained mixed precision achieved state-of-the-art $65.35\%$ performance on average compared to the baseline $64.30\%$ (\textit{i.e.}, GPTQ). Moreover, based on the findings, we introduce novel data-driven techniques for optimizing bit allocation in MoE quantization, including the outlier-aware linear layer scorer and MoE block importance predictor. Our code for reproducing all our experiments is provided at \href{https://github.com/UNITES-Lab/moe-quantization}{https://github.com/UNITES-Lab/moe-quantization}.
\end{abstract}



\received{17 February 2025}

\maketitle

\begin{figure}[t!]
  \centering
  \includegraphics[width=0.99\linewidth]{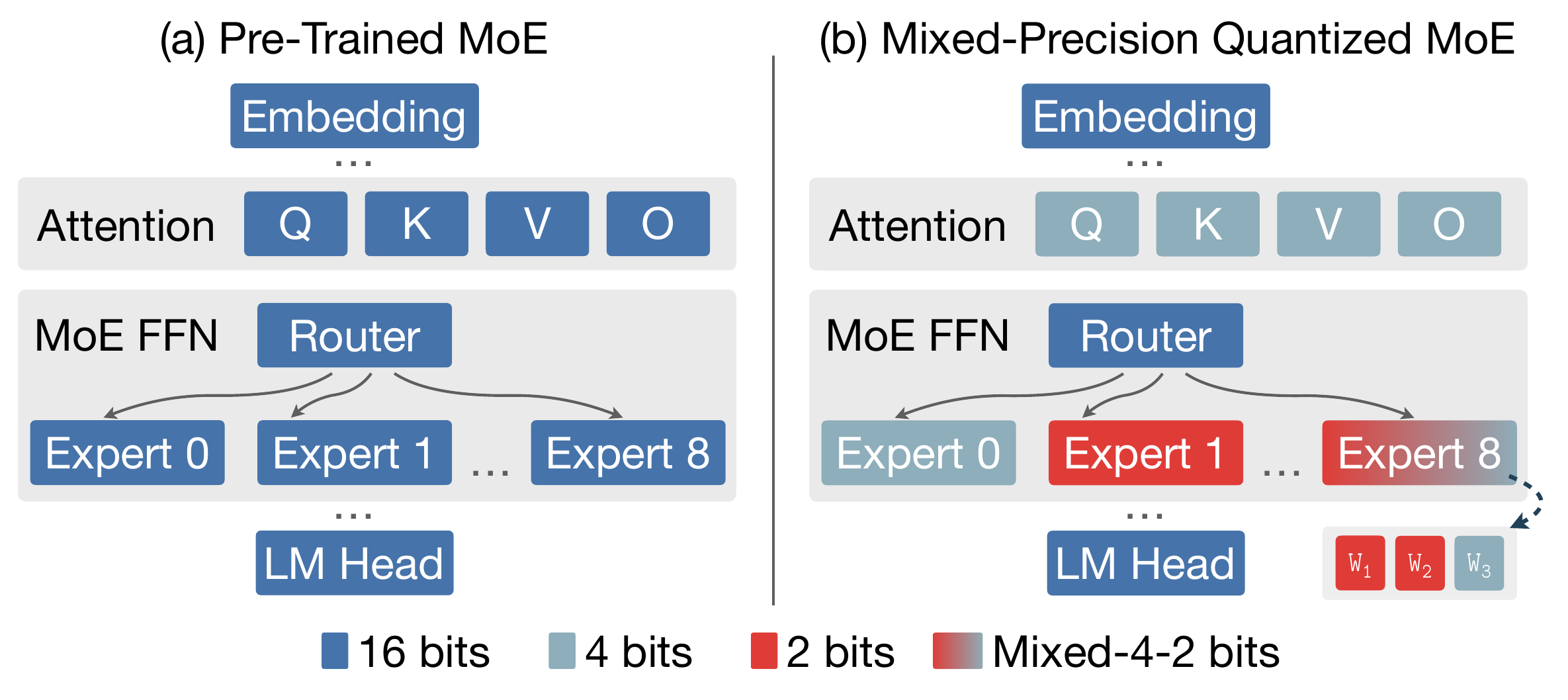}
  \vspace{-3mm}
  \caption{After our post-training quantization, the pre-trained MoE~(a) is quantized into (b) with a mixed-precision based on structures, demonstrated with Mixtral-8x7B.}
  \label{fig:teaser}
  \vspace{-3mm}
\end{figure}

\section{Introduction}

Large Language Models~(LLMs) have achieved remarkable success in various natural language processing tasks, such as language understanding, reasoning, and generation, demonstrating superior performance and adaptability~\citep{brown2020language,jiang2023mistral,kaplan2020scaling,openai2024gpt4,touvron2023llama}. 
However, the rapid growth in model size, with state-of-the-art LLMs containing billions of parameters, poses significant challenges to computational resources and memory consumption~\citep{aminabadi2022deepspeed,lin2024awq,shoeybi2020megatronlm}.
The Mixture of Experts~(MoE)~\citep{shazeer2017outrageously} architecture has emerged as a promising solution to address the computation overhead. 
MoE allows for the scaling up of LLMs while maintaining roughly constant FLOPs. 
By incorporating multiple experts and employing a sparse gating mechanism, MoE achieves efficient computation, enabling the development of larger models within the constraints of limited computation~\citep{dai2024deepseekmoe}.

Despite its advantages, MoE still suffers from extensive memory costs due to its vast parameter size, which hinders its practical deployment and widespread adoption. For example, pretrained in 8-bit precision, the DeepSeek-V3~\citep{liu2024deepseek} MoE model takes around $1.3$ TB memory while only $74$ GB parameters are activated for each input token.
Model compression techniques tailored to MoE architectures are essential to address this issue. 
Existing MoE compression methods can be categorized into two main approaches: merging and pruning.
Expert merging, such as MC-MoE\citep{li2024merge}, aims to reduce the memory footprint by combining similar experts based on routing policy and compressing the resulting model using low-rank decomposition. 
On the other hand, expert pruning, such as task-specific pruning~\citep{chen2022taskspecific}, focuses on identifying and removing the least important experts or connections based on their contribution to a specific task.
However, these approaches not only necessitate prohibitively expensive model retraining but also operate under task-specific settings, limiting their practicality across real-world applications.

Post-training quantization is a family of model compression techniques that converts pre-trained model weights from high-precision formats (\textit{e.g.}, FP$32$) to lower-precision representations without model retraining or task-specific tuning. 
Recent works, such as GPTQ~\citep{frantar2023gptq}, which adapts quantization intervals based on the Hessian information, and SmoothQuant~\citep{lin2024awq}, which jointly quantizes the model weight and activation by offline migrating the activation outliers, have demonstrated the effectiveness of post-training quantization for dense LLMs toward $4$ bits compression. 
This advantage is particularly desired for MoE-based LLMs given their high training cost due to vast parameter sizes and prevailing deployment for various tasks~\cite{jiang2024mixtral,liu2024deepseek}.

However, our experiments show that directly deploy those techniques for MoE models leads to subpar performance. Existing methods employ a uniform bit-width across all components of the MoE model. 
This one-size-fits-all approach fails to account for the inherent sparse structure of the MoE architecture.
For example, the sparse expert activations in MoE models exhibit distinct statistical properties from dense activations, suggesting adaptive bit allocation among experts' quantization. 
This yields the primary \textbf{research question} to be explored:

\textit{(RQ) Do different MoE structures require varying numbers of bits for effective quantization?}

Our investigations reveal that different components in MoE require varying bit allocations, as shown in Figure~\ref{fig:teaser}. For example, shared experts and the first few MoE layers demand higher precision for effective quantization. Moreover, these findings naturally motivate two key questions: ($1$)~How to identify the layers that are more sensitive to quantization; ($2$)~How to systematically determine the importance of each MoE layer for bit allocation. To address these questions, we introduce novel data-driven techniques for optimizing bit allocation in MoE quantization, including the outlier-aware linear layer scorer that captures weight magnitude variations, and the MoE block importance predictor that leverages block-level activations patterns.
Our key contributions are listed:
\begin{enumerate}[leftmargin=*]
    \item We establish the first benchmark for Mixture-of-Experts post-training quantization, \textit{i.e.}, \texttt{QuantMoE-Bench}. This benchmark encompasses investigations into four critical MoE-related heuristics by evaluating different quantization methods including GPTQ and SmoothQuant, and analyzes multiple bit allocation strategies on attention layers, FFNN layers, experts, and MoE blocks. Our evaluation covers two representative MoE LLMs and six benchmark tasks.
    \item Our benchmark study uncovers critical MoE quantization principles: attention layers require higher precision than FFNNs, shared experts need more bits than token-conditioned experts ($4$-bit \textit{v.s.} $2$-bit), and earlier MoE layers demand higher precision compared to later ones. These insights enable optimal bit allocation under constrained memory budgets while maintaining model performance. 
    \item Through extensive experiments, we demonstrate that our fine-grained mixed precision quantization approach achieves state-of-the-art performance, improving average task performance by $1.05\%$ compared to existing methods (\textit{i.e.}, GPTQ).
    \item Leveraging the insights from our benchmark study, we introduce novel data-driven techniques to optimize bit allocation in MoE quantization. These include the development of outlier-aware linear layer scorer and MoE block importance predictor, which significantly improve the effectiveness of mixed-precision quantization by $0.97\%$.
\end{enumerate}

\vspace{-3mm}
\section{Related Works}
\begin{figure*}[t]
    \vspace*{0pt}
    \centering
    \includegraphics[width=0.97\textwidth]{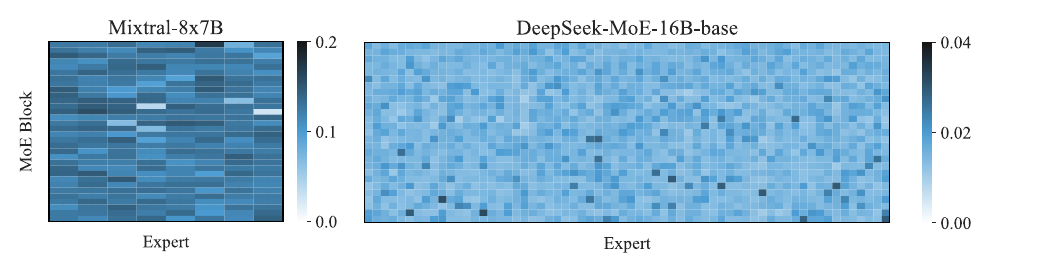}
    \vspace{-10pt}
    \caption{Visualization of expert usage of the two MoE models used in this work. It is profiled on our quantization calibration data, \textit{i.e.}, $512$ random $4096$ token sequences from the WikiText dataset~\citep{merity2016pointer}.}
    \vspace{-10pt}
    \label{fig:routing-frequency}
\end{figure*}

\noindent\textit{\textbf{Mixture-of-Experts.}} The Mixture-of-Experts (MoE) approach~\citep{shazeer2017outrageously} enhances neural network scalability by using router networks to activate model segments according to input tokens selectively. As the dominant architecture in NLP, numerous efforts have adapted feed-forward neural networks~(FFNNs) within Transformers to incorporate MoE layers, constructing MoE language models\citep{dai2024deepseekmoe,fedus2022switch,jiang2024mixtral}. Additionally, several variants of the standard MoE architecture exist. For example, DeepSeek-MoE~\citep{dai2024deepseekmoe} employs numerous finely segmented experts and designates a select few as shared experts to capture common knowledge. MoE's application in LLMs is widely acknowledged for its superior generative abilities and remarkable computing efficiency~\citep{artetxe2022efficient,dai2024deepseekmoe,fedus2022switch,jiang2024mixtral,krajewski2024scaling,rajbhandari2022deepspeedmoe}. The recent work Mixtral~\citep{jiang2024mixtral} illustrates that MoE can match the performance of equivalent full-parameter LLMs while utilizing far fewer active parameters. 
However, MoE suffers from significant memory overhead issues, posing challenges to its efficient deployment~\citep{li2024merge,liu2024deepseek, xue2024moeinfinityoffloadingefficientmoemodel,krajewski2024scaling,luo2024hexamoeefficientheterogeneousawaremoe}.

\noindent\textit{\textbf{MoE Compression.}} MoE models benefit from reduced FLOPs but are constrained by their significant memory overhead. Current works to reduce the memory overhead of MoE models mainly focus on reducing the number of experts. An earlier approach~\citep{chen2022taskspecific} involves pruning non-essential experts for a specific downstream task during fine-tuning, utilizing statistics based on cumulative usage frequency. Another method, MC-SMoE~\citep{li2024merge}, introduces a pipeline that identifies and groups similar experts, subsequently merging them and further decomposing the merged expert into low-rank components within each group. However, these approaches are developed under task-specific fine-tuning settings and do not explore the development of the MoE compression towards a general large langauge model.

\noindent\textit{\textbf{Post-Training Quantization.}} 
Post-training quantization reduces computational and storage demands by converting pre-trained models from high-precision to lower-precision formats without extensive retraining~\citep{frantar2023optimal, frantar2023gptq}. It has been widely applied to LLMs, optimizing them for deployment on resource-constrained devices. Techniques like layer-wise quantization and mixed-precision schemes are designed for minimal performance degradation while reducing model size and computational requirements efficiently~\citep{liu2023llm, pan2023smoothquant+,sharify2024combining}. Recent methods such as SmoothQuant~\citep{xiao2024smoothquant}, GPTQ~\citep{frantar2023gptq}, AWQ~\citep{lin2024awq}, and address specific challenges for LLMs. 
SmoothQuant~\citep{xiao2024smoothquant} ensures smooth precision transitions across layers, reducing quantization errors and maintaining performance.
GPTQ~\citep{frantar2023gptq} employs layer-wise and mixed-precision quantization to balance efficiency and accuracy. 
AWQ~\citep{lin2024awq} adapts to weight sensitivity, preserving critical weights' precision while aggressively quantizing less sensitive ones. 
These advancements in PTQ enable significant reductions in computing and storage while preserving LLM performance. However, their effectiveness on sparse MoE models is under-explored.

\section{Preliminary}

\subsection{Quantization Method}
The primary objective of this work is to benchmark several MoE-related heuristics combined with established LLM quantization techniques. Given that the substantial memory overhead of MoE models predominantly originates from their weights, we adopt GPTQ~\citep{frantar2023gptq}, a popular weight quantization method. GPTQ executes layer-by-layer weight quantization by addressing a specific reconstruction problem for each layer. Specifically, let $\mathbf{W}$ be the weights of a linear layer and $\mathbf{X}$ be the input to that layer derived from a small subset of calibration data, the reconstruction problem is defined as:
\begin{equation}
\texttt{argmin}_{\mathbf{\widehat{W}}} , ||\mathbf{W} \mathbf{X} - \mathbf{\widehat{W}} \mathbf{X}||_2^2.
\end{equation}
This objective, being the sum of squared errors, forms a quadratic equation, allowing the greedy-optimal update of weights to be calculated element-by-element using the Hessian information, $\mathbf{H} = 2\mathbf{X}\mathbf{X}^\top$. GPTQ further enhances this process by incorporating a lazy-batch update and a Cholesky reformulation, to improve scalability and numerical stability for LLM quantization.

\subsection{Mixture-of-Experts}
There are several variants of MoE in the context of LLMs, such as attention MoE and FFNN MoE. In this work, we explore the quantization of MoE models that utilize router networks to selectively activate FFNNs for different input tokens. Specifically, for the $i$-th expert's feed-forward function at the $l$-th transformer layer, denoted as $\mathtt{FFNN}i^l(\cdot)$, the output of the MoE layer for the input hidden states $\mathbf{X}$ is given by:
\begin{equation}
    \mathtt{FFNN}_{\text{MoE}}^l(\mathbf{X}) = \sum_{i=1}^{l} \mathcal{G}(\mathbf{W}_{l}\mathbf{X})\cdot \mathtt{FFNN}_i^l(\mathbf{X}),
\end{equation}
where $\mathbf{W}_l$ represents a linear routing matrix and $\mathcal{G}(\cdot)$ is a routing function that typically employs a top-$k$ selection mechanism, resulting in a sparse output. Due to the duplication of FFNN layers, the principal memory overhead in the MoE model is attributed to the FFNN component.

\subsection{Expert Usage as A Heuristic}\label{sec:usage-heuristic}
As the routing of experts in MoE models is not ideally balanced, expert usage frequency and its variants have emerged as prevalent heuristics for measuring the importance of different experts within an MoE block~\citep{chen2022taskspecific, li2024merge}. For instance, task-specific expert pruning proposed by~\citep{chen2022taskspecific} uses a criterion based on cumulatively calculated expert routing probabilities for pruning during fine-tuning on a specific task. In this paper, focusing on post-training quantization, we utilize the routing distribution from the calibration data as the heuristic for expert usage. Specifically, for the $l$-th MoE block, equipped with a routing matrix $\mathbf{W}_l \in \mathbb{R}^{e\times d}$ and input hidden states $\mathbf{X} \in \mathbb{R}^{b\times d}$ from the calibration data, the expert usage heuristic is:
\begin{equation}
\texttt{usage} = \texttt{normalize}\left(\sum_i\mathcal{G}(\mathbf{W}_{l}\mathbf{X}_{i})\right),
\end{equation}
where $\mathcal{G}(\cdot)$ is the routing function employing a top-$k$ selection mechanism that yields a sparse binary output. We visualize the calculated expert usage of Mixtral-8x7B and DeepSeek-MoE-16B-base MoE models on the quantization calibration data, as shown in Figure~\ref{fig:routing-frequency}. Note that Mixtral-8x7B demonstrates a more balanced routing distribution than DeepSeek-MoE-16B-base.

\section{Benchmark Post-Quantizations for MoE}
In this section, we present several heuristics for MoE quantization and the empirical performance of them. Our benchmarking covers two MoE models and six popular tasks.

\subsection{Benchmark Setups}
\noindent\textit{\textbf{MoE Models.}} We select two representative MoE models for our benchmark evaluation, \textit{i.e.}, Mixtral-8x7B~\citep{jiang2024mixtral} and DeepSeek-MoE-16B-base~\citep{dai2024deepseekmoe}. Mixtral-8x7B substitutes every FFNN with a MoE block and has $8$ experts per MoE block with top-$2$ routing, while DeepSeek-MoE-16B-base uses a fine-grained MoE architecture by including $64$ experts with top-$6$ routing and $2$ shared experts per MoE block. Notably, the DeepSeek-MoE-16B-base model applies a dense FFNN in its first transformer layer while employing an MoE architecture in subsequent blocks for better training stability.

\noindent\textit{\textbf{Quantization.}}
We mainly focus on \textit{weight-only grouped mixed-precision} quantization, though we also extend our experiments and conclusions to its combination with activation quantization in Section~\ref{sec:extended-study}. 
We use GPTQ~\citep{frantar2023gptq} for quantization, without loss of generality.
Throughout this work, we use a group size of $128$. Our experiments emphasize an extreme quantization scenario, where most weights are quantized to either $2$ or $4$ bits.

\noindent\textit{\textbf{Calibration and Evaluation Details.}} We use the calibration data consisting of $512$ random $4096$ token
sequences from the WikiText dataset~\citep{merity2016pointer}, following GPTQ~\citep{frantar2023gptq}. Unlike previous literature that focuses on language modeling benchmarks~\citep{xiao2024smoothquant,lin2024awq,frantar2023gptq}, we evaluate all the methods on six popular LLM tasks for a practical benchmarking: WinoGrande~\citep{ai2:winogrande}, COPA~\citep{gordon-etal-2012-semeval}, OpenBookQA~(OBQA)~\citep{mihaylov2018suit}, HellaSwag~\citep{zellers2019hellaswag}, and MMLU~\citep{hendrycks2021measuring}. We report the performance on MMLU with $5$-shot and all others with zero-shot. All experiments are conducted with PyTorch on $8$ NVIDIA H$100$s, and we utilize \textit{lm-evaluation-harness}~\footnote{https://github.com/EleutherAI/lm-evaluation-harness} for the evaluation of all tasks.

\subsection{Problem Formulation: Pareto-Optimal Bit Allocation}

In this section, we formalize the intrinsic tradeoff between model performance and bit allocation in MoE quantization through the lens of Pareto optimization.

Given an MoE model with $n$ weight components (attention layers, MoE FFNN experts, \textit{etc.}), we aim to find the optimal bit allocation strategy that balances model performance and memory efficiency. Let $\mathbf{b} = (b_1, b_2, \ldots, b_n)$ represent a bit allocation vector, where $b_i \in \{2, 4\}$ is the quantization precision for the $i$-th component. We formulate this as a bi-objective optimization problem:
\begin{equation}
\begin{aligned}
\max_{\mathbf{b}} \quad & \mathcal{P}(\mathbf{b})\\
\vspace{-10pt}
\min_{\mathbf{b}} \quad & \mathcal{B}(\mathbf{b}) = \frac{1}{n}\sum_{i=1}^{n} b_i \\
\text{subject to} \quad & b_i \in \{2, 4\}, \quad \forall i \in \{1, 2, \ldots, n\},
\end{aligned}
\end{equation}
where $\mathcal{P}(\mathbf{b})$ is the model performance and $\mathcal{B}(\mathbf{b})$ is the average bit-width across all components. In practice, we can reformulate this as a constrained optimization problem:
\begin{equation}
\begin{aligned}
\max_{\mathbf{b}} \quad & \mathcal{P}(\mathbf{b}) \\
\text{subject to} \quad & \mathcal{M}(\mathbf{b}) \leq M_{\text{budget}} \\
& b_i \in \mathcal{B}, \quad \forall i \in \{1, 2, \ldots, N\},
\end{aligned}
\end{equation}
where $M_{\text{budget}}$ is the maximum allowed memory budget. This formulation allows us to find the bit allocation that maximizes performance while satisfying memory constraints.

\begin{figure}[t]
  \centering
  \includegraphics[width=0.7\linewidth]{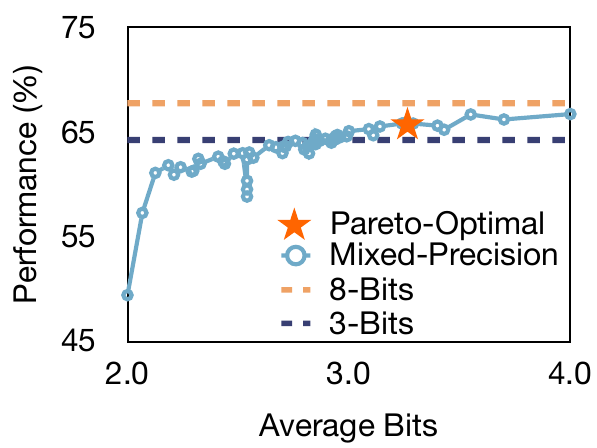}
  \vspace{-13pt}
  \caption{Comparison of allocating more bits~(\textit{i.e.} $4$ bits) for attention and frequent experts with uniform-bits quantization~(\textit{i.e.} $3$ and $8$ bits). The Pareto-optimal solution is $3.29$ bits.}
  \label{fig:pareto-optimal}
  \vspace{-20pt}
\end{figure}

Given the combinatorial nature of this problem and the difficulty in obtaining a closed-form expression for $\mathcal{P}(\mathbf{b})$, we approach it using the structure-aware heuristics described in the following sections, leveraging our observations about the varying importance of different MoE components. As shown in Figure~\ref{fig:pareto-optimal}, our mixed-precision approach effectively navigates the Pareto frontier, offering superior performance compared to uniform bit allocation.

\subsection{Benchmark Results}\label{sec:benchmark-results}

We first evaluate the four bit-allocation heuristic MoE quantization approaches using GPTQ on {Mixtral-8x7B} and {DeepSeek-MoE-16B}. 


\begin{table*}[ht]
\centering
\caption{Comparison of quantization bit allocation strategies including attention (\texttt{Attn}) and shared expert~(\texttt{Shared}) prioritization, frequency-based expert selection~(\texttt{Freq}), first-layer prioritization~(\texttt{FirstL}), linear outlier score predictor~(\texttt{LinearOSP}) and layer importance score predictor~(\texttt{LayerISP}).
We evaluate performance on Mixtral-8x7B and DeepSeek-MoE-16B-base across multiple benchmarks. Unlike uniform-precision quantization~(\textit{i.e.} GPTQ), our approach tailors bit allocation to the MoE structure, demonstrating the effectiveness of structure-aware precision allocation, especially in low-bit settings.} 
\label{tab:main}
\vspace{-3mm}
\resizebox{\linewidth}{!}{
\begin{tabular}{l|c|ccccccc}
\toprule[1pt]
\midrule
   Methodology & Bits  & WinoGrande (\%) & COPA (\%) & OBQA (\%) & HellaSwag (\%) & PIQA (\%) & MMLU (\%) & Average (\%) \\ \midrule
\multicolumn{9}{c}{\texttt{Mixtral-8x7B}} \\ \midrule
GPTQ & $3$ & $73.56$ & $93.00$ & $44.80$ & $75.31$ & $79.11$ & $58.37$ & $70.69$ \\
\midrule
GPTQ & $2$ & $49.33$ & $63.00$ & $25.40$ & $28.18$ & $52.99$ & $24.29$ & $40.53$ \\
\rowcolor[gray]{0.93} 
\texttt{+Attn} & $2.06$ & $54.14$ & $65.00$ & $26.80$ & $44.87$ & $57.78$ & $27.38$ & $46.00$~\textcolor{myorange}{$_{\uparrow 5.47}$} \\
\rowcolor[gray]{0.93} 
\texttt{+Attn+Freq} & $3.27$ & $73.09$ & $92.00$ & $44.80$ & $77.03$ & $78.29$ & $63.05$ & $71.38$~\textcolor{myorange}{$_{\uparrow 30.85}$} \\
\rowcolor[gray]{0.93} 
\texttt{+Attn+Freq+FirstL} & $3.51$ & $73.32$ & $95.00$ & $44.00$ & $80.06$ & $79.71$ & $64.89$ & $72.83$~\textcolor{myorange}{$_{\uparrow 32.30}$} \\
\rowcolor[gray]{0.93} 
\texttt{+Attn+LinearOSP} & $3.12$ & $72.85$ & $92.00$ & $43.40$ & $78.14$ & $78.40$ & $62.92$ & $71.29$~\textcolor{myorange}{$_{\uparrow 30.76}$} \\
\rowcolor[gray]{0.93} 
\texttt{+Attn+LayerISP} & $3.35$ & $71.19$ & $93.00$ & $44.00$ & $77.03$ & $78.29$ & $63.15$ & $71.11$~\textcolor{myorange}{$_{\uparrow 30.58}$} \\

\midrule
\multicolumn{9}{c}{\texttt{DeepSeek-MoE-16B-base}} \\ \midrule

GPTQ & $3$ & $67.87$ & $87.00$ & $40.40$ & $72.92$ & $78.50$ & $39.12$ & $64.30$ \\
\midrule
GPTQ & $2$ & $53.27$ & $76.00$ & $30.20$ & $45.32$ & $66.53$ & $25.28$ & $49.43$ \\
\rowcolor[gray]{0.93} 
\texttt{+Attn} & $2.06$ & $61.72$ & $83.00$ & $34.00$ & $61.93$ & $73.45$ & $26.53$ & $56.77$~\textcolor{myorange}{$_{\uparrow 7.34}$} \\
\rowcolor[gray]{0.93} 
\texttt{+Attn+Shared} & $2.12$ & $65.67$ & $85.00$ & $39.20$ & $67.84$ & $76.28$ & $35.28$ & $61.55$~\textcolor{myorange}{$_{\uparrow 12.12}$} \\
\rowcolor[gray]{0.93} 
\texttt{+Attn+Shared+Freq} & $3.00$ & $68.82$ & $88.00$ & $42.00$ & $73.32$ & $77.31$ & $41.51$ & $65.16$~\textcolor{myorange}{$_{\uparrow 15.90}$} \\
\rowcolor[gray]{0.93} 
\texttt{+Attn+Shared+Freq+FirstL} & $3.06$ & $68.35$ & $88.00$ & $42.60$ & $73.85$ & $77.69$ & $41.58$ & $65.35$~\textcolor{myorange}{$_{\uparrow 15.92}$} \\
\rowcolor[gray]{0.93} 
\texttt{+Attn+Shared+LinearOSP} & $3.06$ & $67.32$ & $88.00$ & $41.60$ & $72.45$ & $77.48$ & $41.64$ & $64.84$~\textcolor{myorange}{$_{\uparrow 15.41}$} \\
\rowcolor[gray]{0.93} 
\texttt{+Attn+Shared+LayerISP} & $3.06$ & $69.30$ & $87.00$ & $42.00$ & $73.56$ & $77.86$ & $41.86$ & $65.27$~\textcolor{myorange}{$_{\uparrow 15.67}$} \\
 \midrule
\bottomrule[1pt]
\end{tabular}}
\vspace{-3mm}
\end{table*}

Table~\ref{tab:main} presents the overall comparison performance of our mixed-precision MoE quantization strategies \textit{v.s.} baseline~(\textit{i.e.} GPTQ) across multiple benchmarks Mixtral-8x7B and DeepSeek-MoE-16B-base. 
The results highlight that uniform quantization leads to substantial performance degradation, reinforcing the need for structure-aware, fine-grained quantization strategies. 
By incorporating methods such as attention-aware adjustments (\texttt{+Attn}), expert frequency-based bit allocation (\texttt{+Freq}), and prioritization of early MoE layers (\texttt{+FirstL}), importance-based quantization (\texttt{+LinearOSP} and \texttt{+LayerISP}), we observe improvements in model performance, particularly in lower-bit settings.
A key takeaway from the results is that a mixed-precision approach tailored to the MoE structure outperforms fixed-bit quantization. 
In particular, expert usage frequency-based quantization (\texttt{+Freq}) improves performance by dynamically allocating more bits to frequently used experts, though its impact varies depending on the routing balance of different models.
Additionally, prioritizing the first few MoE blocks (\texttt{+FirstL}) yields better results. 
Lastly, importance-based strategies(\texttt{+LinearOSP}, \texttt{+LayerISP}) further refine bit allocation, demonstrating the effectiveness of adaptive quantization at different levels of granularity.

To gain deeper insights, we further systematically analyze four key research questions: ($1$) the effectiveness of expert usage frequency as a quantization heuristic, ($2$) whether attention or FFNN layers deserve more bit precision, ($3$) the importance of the first versus last MoE blocks in quantization, and ($4$) the necessity of allocating more bits to shared experts. 
The following sections provide a detailed discussion of each aspect.

\paragraph{\uline{Q1}: Is expert usage frequency a good quantization heuristic? A: Fairly good.}

Expert usage frequency is a popular heuristic in the compression of MoE models, predicated on the insight that less frequently used experts are likely less crucial. Our experiments, detailed in Table~\ref{tab:usage-frequency}, corroborate its effectiveness as a quantization heuristic for MoE models. In particular, for the DeepSeek-MoE-16B-base model, this heuristic outperforms the strategy of randomly allocating more bits to experts, likely due to the model's unbalanced routing distribution. However, with the Mixtral-8x7B model, where the routing distribution is more balanced, the advantage of using expert usage frequency over random allocation is less significant.

\begin{table*}[ht]
\centering
\caption{Comparison of the expert usage frequency heuristic \textit{v.s.} random allocation. For the \texttt{Mixtral-8x7B} model,  we compare the allocation of $4$ bits to the top-\{$2$, $4$\} most frequently used experts per MoE block against randomly selecting \{$2$, $4$\} experts for the same bit allocation. For the \texttt{DeepSeek-MoE-16B-base} model, we keep shared expert \{$8$\} bits and compare between top-\{$10$, $15$, $20$, $25$\} most frequently used experts against randomly selecting \{$10$, $15$, $20$, $25$\} experts per MoE block. The remaining experts are quantized to $2$ bits, while all attention layers are uniformly quantized to $4$ bits. All random experimental results in the format of $a\pm b$ provide the mean value $a$ and its standard deviation $b$ over $3$ independent trials.} \label{tab:usage-frequency}
\vspace{-10pt}
\resizebox{\linewidth}{!}{
\begin{tabular}{l|c|ccccccc}
\toprule[1pt]
\midrule
   Methodology & Bits  & WinoGrande (\%) & COPA (\%) & OBQA (\%) & HellaSwag (\%) & PIQA (\%) & MMLU (\%) & Average (\%) \\ \midrule
\multicolumn{9}{c}{\texttt{Mixtral-8x7B}} \\ \midrule
Random $2$ & $2.54$ & $\mathbf{58.59\pm 2.57}$ & $68.00\pm 11.27$ & $\mathbf{33.00\pm 1.78}$ & $46.60\pm 18.21$ & $60.14\pm 9.32$ & $28.26\pm 4.64$ & $49.10\pm 7.73$ \\
\rowcolor[gray]{0.9} 
Frequent $2$ & $2.54$ & $58.33$ & $\mathbf{76.00}$ & $32.00$ & $\mathbf{56.62}$ & $\mathbf{66.21}$ & $\mathbf{36.01}$ & $\mathbf{54.20}$ \\ \midrule
Random $4$ & $3.03$ & $67.77\pm 0.36$ & $\mathbf{86.33\pm 3.51}$ & $38.47\pm 0.31$ & $67.48\pm 0.52$ & $\mathbf{73.99\pm 0.52}$ & $48.13\pm 2.57$ & $63.70\pm 0.49$ \\
\rowcolor[gray]{0.9} 
Frequent $4$ & $3.03$ & $\mathbf{68.82}$ & $86.00$  & $\mathbf{38.80}$   & $\mathbf{67.68}$   & $72.20$   & $\mathbf{49.42}$  & $\mathbf{63.82}$ \\ \midrule
\multicolumn{9}{c}{\texttt{DeepSeek-MoE-16B-base}} \\ \midrule

Random $10$ & $2.53$ & $\mathbf{67.28}\pm\mathbf{0.04}$ & $\mathbf{88.50}\pm\mathbf{1.50}$ & $38.40\pm0.80$ & $\mathbf{70.99}\pm\mathbf{0.50}$ & $\mathbf{76.74}\pm\mathbf{0.84}$ & $35.23\pm0.09$ & $62.86\pm0.60$\\
\rowcolor[gray]{0.9} 
Frequent $10$ & $2.53$ & $66.46$ & $87.00$    & $\mathbf{39.60}$  & $70.31$ & $76.71$ & $\mathbf{37.84}$ & $\mathbf{62.99}$ \\ \midrule

Random $15$ & $2.68$ & $\mathbf{67.25}\pm\mathbf{0.47}$ & $84.50\pm2.50$ & $\mathbf{40.00}\pm\mathbf{0.60}$ & $\mathbf{71.79}\pm\mathbf{0.43}$ & $76.85\pm0.08$ & $35.71\pm0.82$ & $62.68\pm0.71$\\
\rowcolor[gray]{0.9} 
Frequent $15$ & $2.68$ & $67.17$ & $\mathbf{88.00}$    & $39.00$    & $71.09$ & $\mathbf{76.93}$ & $\mathbf{40.59}$ & $\mathbf{63.80}$\\ \midrule

Random $20$ & $2.83$ & $67.25\pm0.47$ & $84.50\pm2.50$ & $40.00\pm0.60$ & $71.79\pm0.43$ & $76.85\pm0.08$ & $35.71\pm0.82$ & $62.68\pm0.71$ \\
\rowcolor[gray]{0.9} 
Frequent $20$ & $2.83$ & $\mathbf{67.25}$ & $\mathbf{86.00}$ & $\mathbf{40.40}$ & $\mathbf{72.06}$ & $\mathbf{77.58}$ & $\mathbf{40.78}$ & $\mathbf{64.01}$ 
 \\ \midrule

Random $25$ & $2.97$ & $67.72\pm0.24$ & $89.00\pm1.00$ & $\mathbf{40.70}\pm\mathbf{0.10}$ & $71.98\pm0.19$ & $77.04\pm0.05$ & $36.54\pm1.55$ & $63.83\pm0.04$\\
\rowcolor[gray]{0.9} 
Frequent $25$ & $2.97$ & $\mathbf{67.72}$ & $\mathbf{90.00}$ & $39.20$  & $\mathbf{72.83}$ & $\mathbf{77.15}$ & $\mathbf{41.06}$ & $\mathbf{64.66}$ \\ \midrule

\bottomrule[1pt]
\end{tabular}}
\vspace{-2mm}
\end{table*}

\begin{figure}
  \centering
  \includegraphics[width=0.98\linewidth]{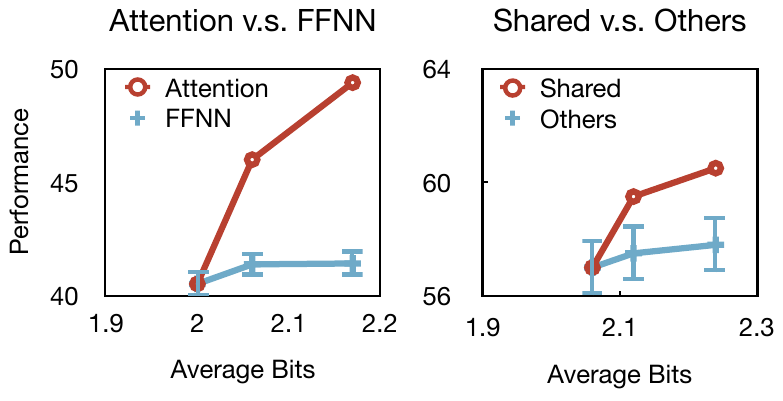}
  \vspace{-10pt}
  \caption{Comparison of quantizing more bits for attention \textit{vs.} FFNN and shared experts \textit{v.s.} others, evaluated on the Mixtral-8x7B model. FFNN and others' results show the mean and standard deviation~(error bars) from $3$ independent trials.}
  \label{fig:attention-vs-ffnn}
  \vspace{-5mm}
\end{figure}

\paragraph{\uline{Q2}: Attention \textit{vs.} FFNN: Which Deserves More Bits in MoE? A: Attention layers are more bit-efficient.}
Because of the unique characteristics of the feedforward neural network (FFNN) within the mixture of experts (MoE) framework. we explore the attention layer and the feedforward neural network layer, which deserves more bits.
We compare the performance evaluated by quantizing the attention layers with more bits \textit{v.s.} randomly selecting experts in the FFNN layers with more bits, maintaining the same average bits of the entire MoE model for a fair comparison. 
Specifically, we quantize the attention weight or randomly selected FFNN weight to \{$2$, $4$, $8$\} bits, while All other weights are quantized to $2$ bits by default. 
As illustrated in Figure~\ref{fig:attention-vs-ffnn} (left), quantizing attention weights to higher bit levels (\textit{i.e.}, $4$ or $8$ bits) consistently results in significant performance gains (over $5\%$) under each average bit allocation for the MoE model. This greater efficiency likely stems from the fact that attention weights are activated for every token, while FFNN weights only engage with a subset of the input tokens. Consequently, increasing the quantization bits for FFNN weights does not benefit all inputs. Based on these findings, attention weights are quantized to $4$ bits by default in all following experiments.

\begin{table*}[htbp]
\centering
\caption{Comparison between quantizing first $k$ \textit{v.s.} last $k$ MoE blocks with higher~(\textit{i.e.} $4$) bits. All weights in attention layers are quantized to $4$ bits, and the other weights are quantized to $2$ bits. In \texttt{DeepSeek-MoE-16B-base} model, we keep the first block that is dense block as $4$ bits by default. We evaluate $k$ of $4$ and $8$. The higher performance of each comparison pair is marked as \textbf{bold}.} \label{tab:layer-depth-bits}
\vspace{-10pt}
\resizebox{\linewidth}{!}{
\begin{tabular}{l|c|ccccccc}
\toprule[1pt]
\midrule
   Methodology & Bits  & WinoGrande (\%) & COPA (\%) & OBQA (\%) & HellaSwag (\%) & PIQA (\%) & MMLU (\%) & Average (\%) \\ \midrule
\multicolumn{9}{c}{\texttt{Mixtral-8x7B}} \\ 
\midrule
\rowcolor[gray]{0.9} 
First $4$ & $2.30$ & $\mathbf{57.85}$ & $\mathbf{72.00}$ & $\mathbf{32.80}$ & $\mathbf{52.80}$ & $\mathbf{61.59}$ & $\mathbf{29.65}$ & $\mathbf{51.12}$ \\
Last $4$ & $2.30$ & $53.75$ & $60.00$ & $27.80$ & $46.25$ & $58.87$ & $26.56$ & $45.54$ \\
\midrule
\rowcolor[gray]{0.9} 
First $8$ & $2.54$ & $\mathbf{62.11}$ & $\mathbf{85.00}$ & $\mathbf{35.80}$ & $\mathbf{62.72}$ & $\mathbf{67.74}$ & $\mathbf{35.61}$ & $\mathbf{58.16}$ \\
Last $8$ & $2.54$ & $52.09$ & $69.00$ & $29.60$ & $47.87$ & $59.58$ & $26.03$ & $47.36$ \\
\midrule
\multicolumn{9}{c}{\texttt{DeepSeek-MoE-16B-base}} \\ \midrule
\rowcolor[gray]{0.9} 
First $4$ & $2.29$ & $\mathbf{65.27}$ & $\mathbf{85.00}$ & $\mathbf{38.40}$ & $\mathbf{64.42}$ & $\mathbf{72.74}$ & $\mathbf{28.88}$ & $\mathbf{59.12}$ \\
Last $4$ & $2.29$ & $62.90$ & $83.00$ & $36.00$ & $64.41$ & $74.65$ & $27.38$ & $58.06$ \\
\midrule
\rowcolor[gray]{0.9} 
First $8$ & $2.63$ & $\mathbf{64.09}$ & $\mathbf{86.00}$ & $\mathbf{38.75}$ & $\mathbf{67.84}$ & $\mathbf{75.35}$ & $30.12$ & $\mathbf{60.36}$ \\
Last $8$ & $2.63$ & $62.83$ & $83.00$ & $37.80$ & $65.94$ & $75.73$ & $\mathbf{31.00}$ & $59.38$ \\


\midrule
\bottomrule[1pt]
\end{tabular}}
\vspace{-2mm}
\end{table*}

\paragraph{\uline{Q3}: Do the model's first or last MoE blocks deserve more bits in quantization? A: The first MoE blocks.}
As more and more Mixture-of-Experts (MoE) architectures emerge, we investigate which layer of the MoE block is more critical and thus deserves more bits during the quantization process.
As shown in Table~\ref{tab:layer-depth-bits}, we evaluate the performance of allocating more bits to the first $k$ blocks versus the last $k$ blocks in quantization. The results consistently indicate that higher bit quantization of the first few blocks yields better performance, suggesting that we can allocate more bits to the quantization of the first blocks of the model. 
This observation aligns with prior studies that have empirically confirmed the greater importance of the first few Transformer blocks~\citep{dai2024deepseekmoe,ma2023llmpruner}.
\vspace{-1mm}

\paragraph{\uline{Q4}: Does the shared expert always deserve more bits? A: Yes.} 
The DeepSeek-MoE-16B-base model includes two shared experts within each MoE block to obtain 
common knowledge across varying domains and alleviate the parameter redundancy. 
To evaluate their role in quantization, we compare quantizing these two shared experts with more bits \textit{v.s.} randomly selecting two non-shared experts for more bit allocation, maintaining the same average bits for a fair comparison. 
The shared or random non-shared experts are quantized to {$2$, $4$, $8$} bits, while attention weights are set to $4$ bits and all other weights to $2$ bits. 
As depicted in Figure~\ref{fig:attention-vs-ffnn} (right), allocating higher bit levels~(\textit{i.e.}, $4$ or $8$ bits) to shared experts consistently yields superior performance. 
This enhanced efficiency and effectiveness are attributed to the shared experts being activated for every input token, unlike non-shared experts, which only engage with specific subsets of the tokens. Allocating more quantization bits to shared experts thus proves to be both more efficient and effective.

\section{Extended Study to Improve MoE Quantization}\label{sec:extended-study}

In this section, we introduce two novel data-driven techniques aimed at enhancing the effectiveness of identifying crucial components within MoE models for improved quantization performance.

\begin{figure*}[t]
    \centering
    \includegraphics[width=0.98\textwidth]{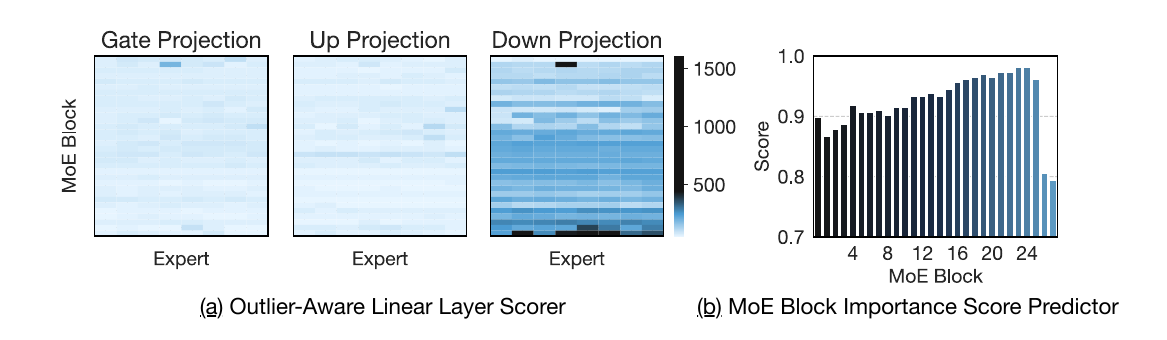}
    \vspace{-16pt}
    \caption{\uline{(a)}~Visualization of the \texttt{outlier-aware linear layer scorer} metric applied to each FFNN linear weight matrix within the Mixtral-8x7B model. For clearer visualization, we present separate components, including the \textit{gate projection}~(left), \textit{up projection}~(middle), and \textit{down projection}~(right) in FFNN experts. \uline{(b)}~Visualization of the \texttt{MoE block importance score predictor} metric applied on the DeepSeek-MoE-16B-base model.}
    \vspace{-10pt}
    \label{fig:score}
\end{figure*}

\subsection{Outlier-Aware Linear Layer Scorer}

\noindent\textit{\textbf{Insight.}} From the quantization perspective, the larger the range of a weight magnitude group, the more difficult it will be for quantization. 
We found that, in MoE, each FFNN linear weight matrix consists predominantly of values within a narrow range, interspersed with a few significant outliers. Consequently, we propose a weight-magnitude-based metric to identify those linear layers that are challenging to quantize effectively, thereby necessitating a higher allocation of quantization bits.

\noindent\textit{\textbf{Methodology.}} We define the metrics to estimate the outliers of weights by the maximum ratio of the largest to the average absolute magnitude within each column. Specifically, for a weight matrix $\mathbf{W} \in \mathbb{R}^{m\times n}$, we compute the metric $\texttt{outlier-score}(\mathbf{W})$ as follows:
\begin{equation}
\texttt{outlier-score}(\mathbf{W}) = \texttt{max}_{j}\left(\frac{\texttt{max}(|\mathbf{W}{:,j}|)}{\texttt{mean}(|\mathbf{W}{:,j}|)}\right),
\end{equation}
where $|\mathbf{W}{:,j}|$ is the absolute value of $\mathbf{W}$'s $j$-th column. With this metric, we can identify those linear layers that require more quantization bits and allocate more to them, providing an effective trade-off between performance and efficiency. The overall procedure is detailed in Algorithm~\ref{alg:outlier-score}.

\begin{algorithm}[htb]
\caption{The Procedure of MoE Mixed-Precision Quantization with \texttt{outlier-score}.}
\begin{algorithmic}[1]
\State \textbf{Initialize:} A MoE model with $l$ linear layers across all the FFNN experts, the number of linear layers for $4$ bit quantization $k$.
\State Let $\mathcal{M}$ and $\mathcal{S}$ represent the set of each linear layer matrix in FFNN and its score, respectively.
\For{linear layer $i=1,\ldots,l$}
    \State $\mathbf{W} \gets \mathcal{M}[i]$
    \State $\mathcal{S}[i] \gets \texttt{max}_{j}\left(\frac{\texttt{max}(|\mathbf{W}{:,j}|)}{\texttt{mean}(|\mathbf{W}{:,j}|)}\right)$    
\EndFor
\State $\alpha \gets \texttt{sorted}(\mathcal{S})[k]$
\State $\mathtt{4}\texttt{bits-quantize}\left(\{\mathcal{M}[i]\mid \mathcal{S}[i] >= \alpha\}\right)$
\State $\mathtt{2}\texttt{bits-quantize}\left(\{\mathcal{M}[i]\mid \mathcal{S}[i] < \alpha\}\right)$
\State \textbf{Return:} A quantized mixed-precision MoE model.
\end{algorithmic}
\label{alg:outlier-score}
\end{algorithm}

\noindent\textit{\textbf{Experiments.}} We evaluate this metric by comparing its application for the top-$p\%$ of linear layers against randomly selecting linear layers, using percentages of $25\%$ and $50\%$. 
In DeepSeek-MoE-16B-base model, we also involve shared experts using this metric.
As illustrated in Table~\ref{tab:linear-layer-outliers}, our proposed scorer consistently outperforms the random baseline on both models and almost all tasks~(except HellaSwag and MMLU). This is particularly evident in the DeepSeek-MoE-16B-base model, where it achieves an average performance improvement of about $3\%$, aligning with our expectations.

\begin{table*}[ht]
\centering
\caption{Comparison between using our linear weight scorer \textit{vs.} random selection of linear layers for bit allocation in quantization. We evaluate by quantizing $25\%$ of the linear layers across all MoE blocks (\textit{i.e.}, FFNN) to $4$ bits. All attention weights are quantized to $4$ bits, and all other weights are quantized to $2$ bits. 
In each comparison pair, the higher performance is highlighted in \textbf{bold}. All random experimental results in the format of $a\pm b$ provide the mean value $a$ and its standard deviation $b$ over $3$ trials.} \label{tab:linear-layer-outliers}
\vspace{-7pt}
\resizebox{\linewidth}{!}{
\begin{tabular}{l|c|ccccccc}
\toprule[1pt]
\midrule
   Methodology & Bits  & WinoGrande (\%) & COPA (\%) & OBQA (\%) & HellaSwag (\%) & PIQA (\%) & MMLU (\%) & Average (\%) \\ \midrule
\multicolumn{9}{c}{\texttt{Mixtral-8x7B}} \\ \midrule
Random $25\%$ & $2.54$ & $60.74\pm 0.63$                   & $78.67\pm 4.62$  & $34.07\pm 1.63$   & $\mathbf{57.36}\pm 0.53$   & $68.19\pm 0.74$  & $\mathbf{32.49\pm 1.60}$ & $55.25\pm 0.95$ \\
\rowcolor[gray]{0.9}
Ours top-$25\%$ & $2.54$ & $\mathbf{62.19}$ & $\mathbf{83.00}$ & $\mathbf{35.80}$ & $57.04$ & $\mathbf{68.23}$ & $30.95$ & $\mathbf{56.20}$ \\
\midrule
\multicolumn{9}{c}{\texttt{DeepSeek-MoE-16B-base}} \\ 
\midrule
Random $25\%$ &2.54 & $64.04\pm0.78$ & $84.67\pm4.73$ & $37.53\pm0.46$ & $67.39\pm0.71$ & $74.61\pm0.60$ & $29.43\pm1.31$ & $59.61\pm0.76$ \\
\rowcolor[gray]{0.9}
Ours top-$25\%$ & $2.54$ & $\mathbf{66.14}$ & $\mathbf{85.00}$ & $\mathbf{38.80}$ & $\mathbf{71.65}$ & $\mathbf{76.82}$ & $\mathbf{36.19}$ & $\mathbf{62.43}$   \\
\midrule
\bottomrule[1pt]
\end{tabular}}
\end{table*}

\begin{table*}[ht]
\centering
\caption{Comparison between using our MoE block importance predictor \textit{v.s.} two baselines: \ding{172}random selecting and \ding{173}first $k$ MoE blocks. The predicted or selected MoE blocks are quantized to $4$ bits, all attention weights are quantized to $4$ bits, and all other weights are quantized to $2$ bits. In each comparison, the highest performance is highlighted in \textbf{bold}. All random experimental results in the format of $a\pm b$ provide the mean value $a$ and its standard deviation $b$ over $3$ independent trials.} \label{tab:block-predictor}
\vspace{-7pt}
\resizebox{\linewidth}{!}{
\begin{tabular}{l|c|ccccccc}
\toprule[1pt]
\midrule
   Methodology & Bits  & WinoGrande (\%) & COPA (\%) & OBQA (\%) & HellaSwag (\%) & PIQA (\%) & MMLU (\%) & Average (\%) \\ \midrule
\multicolumn{9}{c}{\texttt{DeepSeek-MoE-16B-base}} \\ \midrule
Random $4$ & $2.29$ & $61.09\pm0.78$ & $83.00\pm0.00$ & $37.20\pm0.85$ & $64.88\pm0.30$ & $74.21\pm0.08$ & $27.82\pm0.46$ & $58.03\pm0.13$\\

First $4$ & $2.29$ & $65.27$ & $\mathbf{85.00}$ & $\mathbf{38.40}$ & $64.42$ & $72.74$ & $28.88$ & $59.12$ \\
\rowcolor[gray]{0.9}
Predicted $4$ & $2.29$ & $\mathbf{65.27}$ & $83.00$ & $36.60$ & $\mathbf{64.88}$ & $\mathbf{74.54}$ & $\mathbf{37.75}$ & $\mathbf{60.34}$ 
\\\midrule
Random $8$ & 2.63 & $64.48\pm0.83$ & $85.33\pm3.21$ & $38.73\pm0.95$ & $67.57\pm0.40$ & $\mathbf{75.43}\pm\mathbf{0.14}$ & $\mathbf{31.41}\pm\mathbf{2.17}$ & $60.49\pm0.56$\\
First $8$ & $2.63$ &  $64.09$ & $\mathbf{86.00}$ & $\mathbf{38.75}$ & $67.84$ & $75.35$ & $30.12$ & $60.36$ \\
\rowcolor[gray]{0.9} 
Predicted $8$ & $2.63$ & $\mathbf{65.35}$ & $\mathbf{86.00}$ & $38.00$ & $\mathbf{68.77}$ & $75.35$ & $30.01$ & $\mathbf{60.58}$  
\\\midrule
Random $12$ & 2.92 & $64.64\pm0.89$ & $83.50\pm0.71$ & $\mathbf{39.60}\pm2.83$ & $69.51\pm0.56$ & $75.98\pm0.42$ & $32.57\pm0.30$ & $60.97\pm0.62$ \\
First $12$ & $2.92$ & $67.48$ & $\mathbf{88.00}$ & $38.60$ & $70.59$ & $75.95$ & $\mathbf{39.25}$ & $63.31$ \\
\rowcolor[gray]{0.9} 
Predicted $12$ & $2.92$ & $\mathbf{68.11}$ & $\mathbf{88.00}$ & $39.20$ & $\mathbf{71.82}$ & $\mathbf{76.66}$ & $38.45$ & $\mathbf{63.71}$ 
\\\midrule
\bottomrule[1pt]
\end{tabular}}
\vspace{-2mm}
\end{table*}

\noindent\textit{\textbf{Visualization.}} As shown in Figure~\ref{fig:score} (a), we visualize the proposed \texttt{outlier-score} for each FFNN linear weight within the Mixtral-8x7B model. Given that each FFNN expert includes three linear layers, namely the \textit{gate projection}, \textit{up projection}, and \textit{down projection}, we visualize these components separately to ensure clarity. Notably, many of the \textit{down projection} linear layers, particularly those positioned later in the MoE model, exhibit significantly higher \texttt{outlier-scores} compared to others.

\subsection{MoE Block Importance Score Predictor}

\begin{algorithm}[htb]
\caption{The Training Procedure of Block Score Predictor.}
\begin{algorithmic}[1]
\State \textbf{Initialize:} A MoE block $M$, token input and output embedding set at block $M$ $\{(\mathbf{x}_i, \mathbf{y}_i)\}_{i\in [N]}$.
\State Let $\mathcal{BSP}$ denotes the block score predictor.
\State $\mathcal{X} \gets \left\{\mathbf{x}_i \mid i\in [N]\right\}$
\State $\mathcal{S} \gets \left\{\texttt{cosine}(\mathbf{x}_i, \mathbf{y}_i) \mid i\in [N]\right\}$
\State $\mathcal{BSP} \gets \texttt{train}(\mathcal{X}, \mathcal{S})$
\State \textbf{Return:} The importance scorer $\mathcal{BSP}$ for MoE Block $M$.
\end{algorithmic}
\label{alg:block-predictor}
\end{algorithm}

Inspired by \uline{Q3} in Section~\ref{sec:benchmark-results}, which demonstrates that allocating more bits to different MoE blocks yields variable performance improvements, we propose a novel method to identify and quantize those critical blocks with additional bits. Specifically, this section outlines our approach to calculating importance scores for bit allocation using a data-driven method with a lightweight predictor.

\noindent\textit{\textbf{Insight.}} We find an increasing cosine similarity between the tensors generated before and after the FFN blocks for some of the MoE blocks, indicating less important computation results produced by these blocks. This observation also aligns with observations on dense models in previous literature~\citep{jaiswal2024ffnskipllm}. Therefore, the basic idea is that less accurate output of these blocks producing tokens with high cosine similarity will not affect the overall model performance much, thus lower weight bits might not hurt performance much. 

\noindent\textit{\textbf{Methodology.}} To capture the generalized hidden states' dynamic information of each MoE block, we train a two-layer FFNN with a tangent activation function. This network predicts the cosine similarity between the input and output hidden states. We utilize a dataset of $400$ random sequences, each containing $1024$ tokens from the WikiText dataset~\citep{merity2016pointer}, for training. The detailed training procedure is in Algorithm~\ref{alg:block-predictor}. During quantization, we applied this predictor to the calibration data, computing an average predicted score for each MoE layer across all tokens. A higher predicted score indicates less importance and fewer bits for quantization.

\noindent\textit{\textbf{Experiments.}} In Table~\ref{tab:block-predictor}, we compare the performance of using our block importance predictor to select $k$ MoE blocks for $4$ bits and others for $2$ bits quantization with two other baselines: \ding{172} random selecting $k$ MoE blocks, and \ding{173} first $k$ MoE blocks~(as it is the best in \uline{Q3} in Section~\ref{sec:benchmark-results}). Evaluation results on the DeepSeek-MoE-16B-base model are presented in Table~\ref{tab:block-predictor}, showing the superiority of our method against the other two baselines.

\noindent\textit{\textbf{Visualization.}} We visualize the predicted scores of each MoE block using our trained predictors in the DeepSeek-MoE-16B-base model, as shown in Figure~\ref{fig:score} (b). Notably, MoE blocks situated in the middle of the model, which exhibit higher scores, are regarded as less critical. Consequently, these blocks will be quantized with fewer bits (specifically, $2$ bits), reflecting their lower importance.
Besides, Figure~\ref{fig:score} (b) also demonstrates that the first few MoE blocks are more important aligned with \uline{Q3}. Interestingly, the last two blocks of the DeepSeek-MoE-16B-base model are also crucial, thereby allocating more bits and yielding better performance.

\vspace{-10pt}
\section{Conclusion}
This work introduces mixed-precision MoE quantization through a systematic investigation of various heuristic-based approaches in the post-training setting. While conventional quantization techniques~(\textit{e.g.}, GPTQ) show limited effectiveness when directly applied to MoE models, the question of optimal bit allocation across different MoE model components requires deeper exploration. We present \texttt{QuantMoE-Bench}, the first comprehensive benchmark to comprehensively study post-training quantization of MoE models, which reveals critical insights, including significant importance variations among MoE blocks. Drawing on these insights, we further develop innovative techniques, \textit{i.e.}, a block importance predictor and a linear layer outlier range scorer, to more precisely identify which components benefit most from higher-precision quantization. Our methods substantially improve the effectiveness of MoE model quantization. We hope our study provides novel insights into the memory compression of MoE-based models.


\bibliographystyle{ACM-Reference-Format}
\bibliography{main}

\appendix
\section{Appendix}

\subsection{Evaluation Datasets}

In this section, we introduce details of the datasets in our evaluation. 
For a more comprehensive study, we have selected six popular benchmark tasks: WinoGrande, COPA, OpenBookQA (OBQA), HellaSwag, and MMLU. 

\textbf{WinoGrande~\cite{ai2:winogrande}} is a large-scale dataset designed for commonsense reasoning, consisting of pronoun resolution problems. Each instance in the dataset presents a sentence with an ambiguous pronoun that needs to be resolved based on context. This task tests the model's ability to understand and reason about everyday situations.

\textbf{The Choice of Plausible Alternatives (COPA) }dataset~\cite{gordon-etal-2012-semeval} focuses on causal reasoning. Each question in COPA consists of a premise and two choices, where the model must select the more plausible alternative. This task evaluates the model's understanding of cause-and-effect relationships in natural language. 

\textbf{OpenBookQA}~\cite{mihaylov2018suit} is a multiple-choice question-answering dataset that requires the model to use both scientific facts and commonsense knowledge. The dataset challenges the model's ability to combine factual knowledge with reasoning to answer questions correctly.

\textbf{HellaSwag}~\cite{zellers2019hellaswag} is a benchmark for commonsense NLI (Natural Language Inference) that tests the model's ability to predict the most plausible continuation of a given sentence. The dataset contains scenarios from various domains, such as cooking and sports, requiring the model to understand context and plausibility.

\textbf{The Massive Multitask Language Understanding (MMLU) }benchmark~\cite{hendrycks2021measuring} evaluates models across a wide range of subjects, from elementary mathematics to law. For this study, we report performance on MMLU with a 5-shot setting, where the model is given five examples per task before evaluation, allowing us to gauge the model's few-shot learning capabilities.

We perform a zero-shot evaluation on WinoGrande, COPA, OpenBookQA, and HellaSwag, where the model is not provided with any task-specific training examples. 
For MMLU, a 5-shot evaluation protocol is adopted, providing five examples per task. 
This setup helps us assess the generalization ability of the models across different types of reasoning and knowledge-based tasks.

\subsection{Random Seed}
For all the random selection experiments, we use random seeds \{$42$, $43$, $44$\} to conduct three independent trials and then report the standard deviation and mean.

\subsection{Further Discussion}
In this section, we present further discussion of the DeepSeek-MoE-16B-base performance across different bits. 

\noindent\textit{\textbf{Expert usage frequency.}}As shown by \uline{Q1} in Section \ref{sec:benchmark-results}, expert usage frequency is a critical metric in the compression of MoE models, predicated on the insight that less frequently used experts are likely less crucial. 
We present further discussion of ablation on the bits allocation in the expert-frequency-based methods.

\begin{table*}[htbp]
  \centering
  \caption{Ablation on the allocated bits for the selected top-$k$ experts based on frequency. We compare the allocation of \{$4$, $8$\} bits of the top-$k$ experts based on frequency, and all other experts are quantized to $2$ bits.}
  \resizebox{0.9\linewidth}{!}{
    \begin{tabular}{ccc|ccccccc}
    \toprule[1pt]
    \midrule
    Top   & Top-$k$ bits & Bits  & WinoGrande (\%) & COPA (\%) & OBQA (\%) & HellaSwag (\%) & PIQA (\%) & MMLU (\%) & Average (\%) \\ 
    \midrule
    \multirow{2}[2]{*}{$1$} & $4$     & $2.29$  & $66.30$ & $83.00$ & $39.00$ & $69.28$ & $75.03$ & $35.02$ & $61.27$ \\
          & $8$     & $2.35$  & $66.14$ & $87.00$ & $39.80$ & $69.44$ & $75.30$ & $34.04$ & $61.95$ \\
    \midrule
    \multirow{2}[2]{*}{$2$} & $4$     & $2.32$  & $66.38$ & $88.00$ & $38.60$ & $69.44$ & $76.06$ & $36.49$ & $62.49$ \\
          & $8$     & $2.44$  & $65.98$ & $90.00$ & $38.60$ & $69.77$ & $76.33$ & $35.82$ & $62.75$ \\
    \midrule
    \multirow{2}[2]{*}{$5$} & $4$     & $2.41$  & $66.54$ & $87.00$ & $38.40$ & $70.13$ & $76.12$ & $38.02$ & $62.70$ \\
          & $8$     & $2.70$  & $64.96$ & $89.00$ & $39.40$ & $70.56$ & $75.90$ & $38.56$ & $63.06$ \\
    \midrule
    \multirow{2}[2]{*}{$10$} & $4$     & $2.55$  & $67.17$ & $86.00$ & $39.20$ & $70.55$ & $76.55$ & $39.11$ & $63.10$ \\
          & $8$     & $3.14$  & $66.06$ & $88.00$ & $39.00$ & $70.81$ & $76.71$ & $39.30$ & $63.31$ \\
    \midrule
    \multirow{2}[1]{*}{$15$} & $4$     & $2.70$  & $67.17$ & $83.00$ & $39.00$ & $71.72$ & $76.93$ & $40.41$ & $63.04$ \\
          & $8$     & $3.58$  & $65.75$ & $85.00$ & $41.00$ & $71.34$ & $76.39$ & $40.48$ & $63.33$ \\
    \midrule
          
    \multirow{2}[1]{*}{$20$} & $4$     & $2.85$  & $67.88$ & $84.00$ & $40.20$ & $72.35$ & $77.69$ & $41.25$ & $63.90$ \\
          & $8$     & $4.02$  & $66.61$ & $89.00$ & $38.00$ & $72.58$ & $77.64$ & $41.25$ & $64.18$ \\
    \midrule
    \multirow{2}[1]{*}{$25$} & $4$     & $2.99$  & $67.17$ & $87.00$ & $40.00$ & $73.26$ & $78.07$ & $42.38$ & $64.65$ \\
          & $8$     & $4.46$  & $68.67$ & $86.00$ & $41.00$ & $73.00$ & $78.67$ & $41.79$ & $64.86$ \\
    \midrule
          
    \multirow{2}[1]{*}{$30$} & $4$     & $3.14$  & $69.69$ & $89.00$ & $40.60$ & $73.92$ & $77.53$ & $42.82$ & $65.59$ \\
          & $8$     & $4.90$  & $67.56$ & $88.00$ & $40.80$ & $73.88$ & $78.56$ & $41.94$ & $65.12$ \\
    \midrule
    \bottomrule[1pt]
    \end{tabular}%
  \label{tab:deepseek-topk}}
\end{table*}%



In Table~\ref{tab:deepseek-topk}, we compare the allocation of \{$4$, $8$\} bits of the selected top-$k$ experts, while all other experts are quantized to $2$ bits. 
We quantize the shared experts and attention weights to $8$ bits. 
Table~\ref{tab:deepseek-topk} indicates that increasing the bit width of frequently activated experts improves performance. 
However, the gain from increasing the top-$k$ expert bits from $4$ to $8$ is minimal.


\begin{table}[htbp]
  \centering
  \setlength{\tabcolsep}{2pt}
\small
  \caption{Baseline results of the $16$-bit (FP$16$), $4$-bit, and $2$-bit quantization.}
  \vspace{5pt}
  \resizebox{\linewidth}{!}{
    \begin{tabular}{c|ccccccc}
    \toprule[1pt]
    \midrule
    Bits  & WinoGrande (\%)  & COPA (\%) & OBQA (\%) & HellaSwag (\%) & PIQA (\%) & MMLU (\%) & Average (\%) \\ 
    \midrule 
    \multicolumn{8}{c}{{Mixtral-8x7B}} \\
    \midrule 
     $16$ & $76.48$ & $93.00$ & $47.00$ & $83.98$ & $82.37$ & $70.35$ & $75.33$  \\
     $4$ & $74.98$ & $92.00$ & $46.20$ & $81.65$ & $80.85$ & $67.65$ & $73.89$  \\
     $2$ & $49.33$ & $63.00$ & $25.40$ & $28.18$ & $52.99$ & $24.29$ & $40.53$  \\
     \midrule
     \multicolumn{8}{c}{{DeepSeek-MoE-16B-base}} \\
    \midrule 
     $16$ & $70.40$ & $91.00$ & $44.20$ & $77.35$ & $78.72$ & $44.77$ & $67.74$  \\
     $4$  & $71.35$ & $87.00$ & $43.20$ & $76.39$ & $78.51$ & $44.22$ & $66.78$  \\
     $2$  & $53.28$ & $76.00$ & $30.20$ & $45.33$ & $66.54$ & $25.28$ & $49.44$  \\
    \midrule
    \bottomrule[1pt]
    \end{tabular}
  \label{tab:baseline}}
\end{table}

\noindent\textit{\textbf{Combination of the weight outlier and expert usage frequency.}} We conducted additional experiments on the DeepSeek-MoE-16B-base model by integrating bit-width allocation based on layers with significant weight outliers with allocation based on expert usage frequency to explore the trade-off between them. Specifically, we aimed for a total average bit budget of $2.97$. We select portions of the model to be quantized to $4$ bits using a combination of the two heuristics, while quantizing all attention weights to $4$ bits and all other weights to $2$ bits. For selecting the $4$-bit weights, we introduce a hyper-parameter, $\alpha$ ($0$ < $\alpha$ < $1$), representing the proportion of weights chosen based on expert usage frequency, with the remainder selected based on weight outliers. We varies $\alpha$ to illustrate the trade-off between these methods, as detailed above. As shown in Table~\ref{tab:combine-outlier-frequency}, the optimal combination of these two methods occurs when alpha is set to $0.1$. This means that $20\%$ of the $4$-bit MoE weights are selected based on expert usage frequency, while the remaining $80\%$ are chosen according to weight outliers.

\begin{table}[htbp]
  \centering
  \caption{The combination of weight outlier and expert usage frequency, evaluated on the DeepSeek-MoE-16B-base model.}
  \label{tab:combine-outlier-frequency}
  \resizebox{\linewidth}{!}{
    \begin{tabular}{cc|cccccccc}
    \toprule[1pt]
    \midrule
    Bits   & $\alpha$ & WinoGrande (\%)  & COPA (\%) & OBQA (\%) & HellaSwag (\%) & PIQA (\%) & MMLU (\%) & Average (\%) \\ 
    \midrule
    \multirow{11}[2]{*}{$2.97$} & $0.0$     & $67.72$  & $90.00$ & $39.20$ & $72.83$ & $77.15$ & $41.06$ & $64.66$  \\
          & $0.1$     & $68.11$  & $89.00$ & $41.60$ & $72.88$ & $77.80$ & $41.84$ & $65.21$ \\
          & $0.2$     & $69.21$  & $89.00$ & $41.20$ & $72.60$ & $76.93$ & $41.60$ & $65.09$ \\
          & $0.3$     & $68.92$  & $88.00$ & $42.00$ & $72.06$ & $76.65$ & $41.21$ & $64.81$  \\
          & $0.4$     & $67.48$  & $89.00$ & $41.40$ & $71.88$ & $76.71$ & $40.96$ & $64.57$  \\
          & $0.5$     & $67.32$  & $90.00$ & $40.80$ & $71.89$ & $76.93$ & $40.21$ & $64.52$  \\
          & $0.6$     & $65.90$  & $87.00$ & $39.40$ & $71.86$ & $76.76$ & $38.67$ & $63.27$  \\
          & $0.7$     & $66.21$  & $87.00$ & $41.40$ & $71.45$ & $76.87$ & $36.98$ & $63.32$  \\
          & $0.8$     & $66.45$  & $89.00$ & $41.00$ & $70.89$ & $76.60$ & $37.67$ & $63.60$  \\
          & $0.9$     & $66.37$  & $84.00$ & $40.20$ & $70.83$ & $76.87$ & $39.84$ & $63.02$  \\
          & $1.0$     & $68.19$  & $87.00$ & $41.60$ & $71.01$ & $76.11$ & $40.81$ & $64.12$  \\
    \midrule
    \bottomrule[1pt]
    \end{tabular}}
\end{table}

\begin{table*}[htbp]
  \centering
  \caption{Baseline results of the $16$-bit (FP$16$), $4$-bit, and $2$-bit quantization.}
  \resizebox{0.9\linewidth}{!}{
    \begin{tabular}{c|ccccccc}
    \toprule[1pt]
    \midrule
    Bits  & WinoGrande (\%)  & COPA (\%) & OBQA (\%) & HellaSwag (\%) & PIQA (\%) & MMLU (\%) & Average (\%) \\ 
    \midrule 
    \multicolumn{8}{c}{Mixtral-8x7B} \\
    \midrule 
     $16$ & $76.48$ & $93.00$ & $47.00$ & $83.98$ & $82.37$ & $70.35$ & $75.33$  \\
     $4$ & $74.98$ & $92.00$ & $46.20$ & $81.65$ & $80.85$ & $67.65$ & $73.89$  \\
     $2$ & $49.33$ & $63.00$ & $25.40$ & $28.18$ & $52.99$ & $24.29$ & $40.53$  \\
     \midrule
     \multicolumn{8}{c}{DeepSeek-MoE-16B-base} \\
    \midrule 
     $16$ & $70.40$ & $91.00$ & $44.20$ & $77.35$ & $78.72$ & $44.77$ & $67.74$  \\
     $4$  & $71.35$ & $87.00$ & $43.20$ & $76.39$ & $78.51$ & $44.22$ & $66.78$  \\
     $2$  & $53.28$ & $76.00$ & $30.20$ & $45.33$ & $66.54$ & $25.28$ & $49.44$  \\
    \midrule
    \bottomrule[1pt]
    \end{tabular}
  \label{tab:baseline}}
\end{table*}

\noindent\textit{\textbf{Baseline results of low-precision quantization.}} We provide the $16$-bit (FP$16$), $4$-bit, and $2$-bit baselines of both Mixtral-8x7B and DeepSeek-MoE-16B-base models in Table~\ref{tab:baseline}.

\end{document}